\documentclass[letterpaper, 10 pt, conference]{ieeeconf}  
\IEEEoverridecommandlockouts
\usepackage{cite}
\usepackage{balance}
\usepackage{comment}
\usepackage[ruled,linesnumbered]{algorithm2e}
\usepackage{amsmath,amssymb,amsfonts}
\usepackage{subfigure}
\usepackage{graphicx} 
\usepackage{textcomp}
\usepackage{multirow} 
\usepackage{xcolor}
\usepackage{float}
\usepackage{tabularray}
\usepackage{hyperref}
\usepackage{array}
\usepackage{booktabs}
\usepackage{verbatim} 
\usepackage{makecell} 
\usepackage{booktabs} 
\usepackage{algpseudocode}
\makeatletter

\newcommand{\Rmnum}[1]{\expandafter\@slowromancap\romannumeral #1@}

\makeatother

\def\BibTeX{{\rm B\kern-.05em{\sc i\kern-.025em b}\kern-.08em
    T\kern-.1667em\lower.7ex\hbox{E}\kern-.125emX}}

\title{\LARGE \bf Sample-Efficient Learning with Online Expert Correction for Autonomous Catheter Steering in Endovascular Bifurcation Navigation}
\author{Hao Wang$^{1}$, Tianliang Yao$^{2}$, Bo Lu$^{3}$, Zhiqiang Pei$^{4}$, Liu Dong$^{5}$, Lei Ma$^{1}$, Peng Qi$^{1, 6, *}$
\thanks{This work has been accepted by IEEE ICRA 2026. Copyright may be transferred, after which this version may no longer be accessible.}
\thanks{This work is supported by the National Key Research and Development Program of China under Grant No. 2023YFB4705200, the National Natural Science Foundation of China under Grant No. 62273257, and the Open Project Fund of State Key Laboratory of Cardiovascular Diseases No.2024SKL-TJ002. \emph{(*Corresponding Author: Peng Qi, email: pqi@tongji.edu.cn)}.}
\thanks{$^{1}$Department of Control Science and Engineering, College of Electronics and Information Engineering, and Shanghai Institute of Intelligent Science and Technology, Tongji University, Shanghai 200092, China;}%
\thanks{$^{2}$Department of Electronic Engineering, Faculty of Engineering, The Chinese University of Hong Kong, Hong Kong SAR 999077, China;}
\thanks{$^{3}$Robotics and Microsystems Center, School of Mechanical and Electric Engineering, Soochow University, Suzhou, Jiangsu 215131, China;}%
\thanks{$^{4}$School of Oriental Pan-Vascular Devices Innovation College, University of Shanghai for Science and Technology, Shanghai 200093, China;}%
\thanks{$^{5}$Shanghai Operation Robot Co., Ltd., Shanghai 201318, China;}
\thanks{$^{6}$State Key Laboratory of Cardiovascular Diseases and Medical Innovation Center, Shanghai East Hospital, School of Medicine, Tongji University 200092, Shanghai, China.}
}

\begin{document}

\maketitle 
\pagestyle{empty}  
\thispagestyle{empty} 

\begin{abstract}
Robot-assisted endovascular intervention offers a safe and effective solution for remote catheter manipulation, reducing radiation exposure while enabling precise navigation. Reinforcement learning (RL) has recently emerged as a promising approach for autonomous catheter steering; however, conventional methods suffer from sparse reward design and reliance on static vascular models, limiting their sample efficiency and generalization to intraoperative variations. To overcome these challenges, this paper introduces a sample-efficient RL framework with online expert correction for autonomous catheter steering in endovascular bifurcation navigation. The proposed framework integrates three key components: (1) A segmentation-based pose estimation module for accurate real-time state feedback, (2) A fuzzy controller for bifurcation-aware orientation adjustment, and (3) A structured reward generator incorporating expert priors to guide policy learning. By leveraging online expert correction, the framework reduces exploration inefficiency and enhances policy robustness in complex vascular structures. Experimental validation on a robotic platform using a transparent vascular phantom demonstrates that the proposed approach achieves convergence in 123 training episodes—a 25.9\% reduction compared to the baseline Soft Actor-Critic (SAC) algorithm—while reducing average positional error to 83.8\% of the baseline. These results indicate that combining sample-efficient RL with online expert correction enables reliable and accurate catheter steering, particularly in anatomically challenging bifurcation scenarios critical for endovascular navigation.
\end{abstract}

\section{Introduction}
\begin{figure}[!htbp]
\centering
\includegraphics[width=0.50\textwidth]{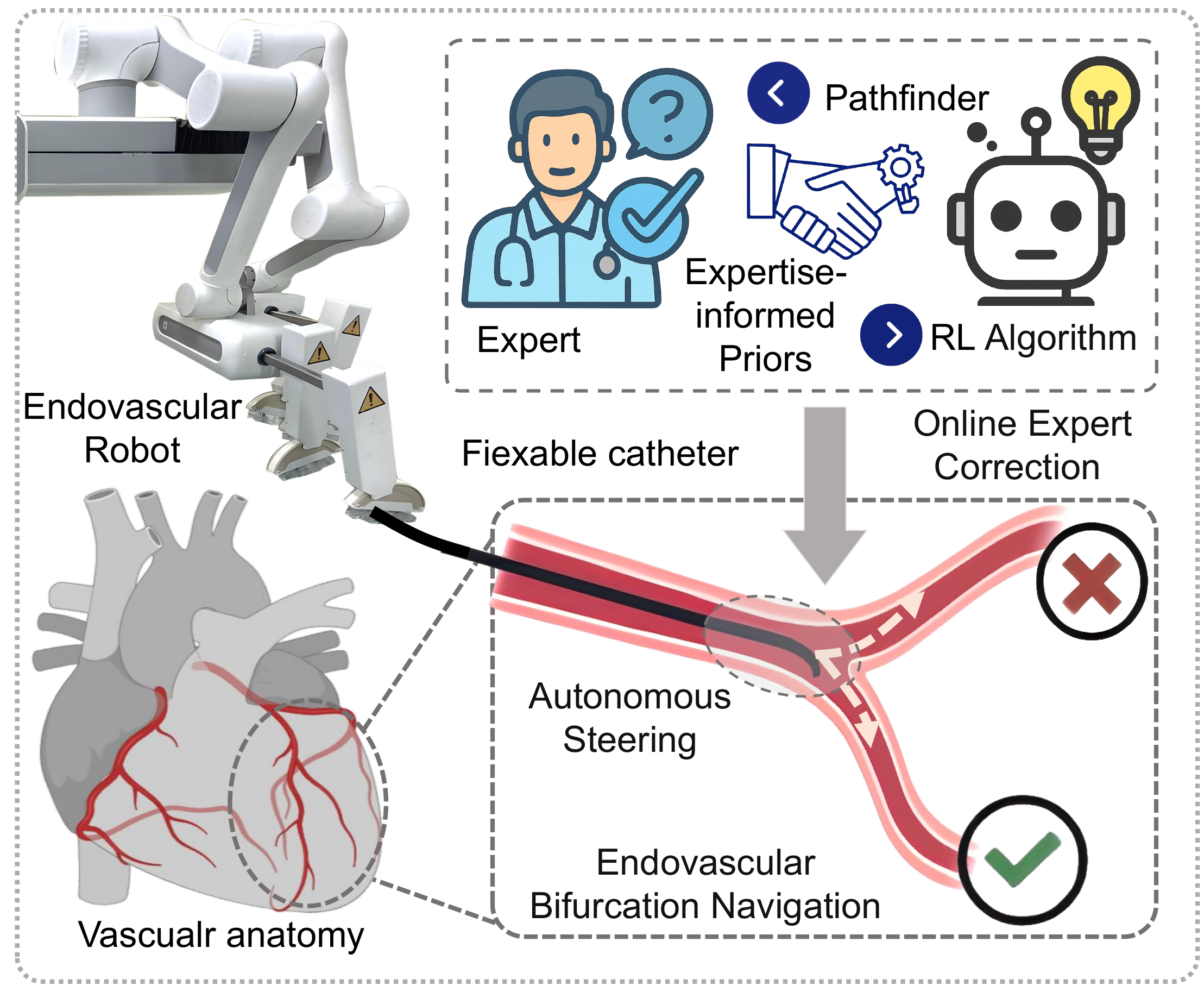}
\caption{Illustration of endovascular navigation enhanced by expert knowledge and behavior modeling. The framework utilizes intraoperative imaging to detect the real-time position and orientation of the catheter tip. By combining expert procedural patterns with reinforcement learning strategies, the system dynamically adjusts navigation to ensure safe and efficient traversal through vascular bifurcations. This hybrid approach improves accuracy and reliability in reaching target anatomical sites during complex interventions. The schematic was created using BioRender (\url{https://biorender.com}).}
\label{fig:Introduction}
\end{figure}

Endovascular interventions are a minimally invasive approach to the diagnosis and treatment of complex cardiovascular diseases, offering shorter recovery times and fewer postoperative complications than open surgery \cite{li2024machine}. Effective catheter navigation underlies procedural safety and efficiency, and bifurcation navigation demands accurate branch selection that depends on precise catheter steering in real-time \cite{konda2025robotically}. Robot-assisted systems offer motion scaling and tremor suppression that enable consistent steering and standardized bifurcation navigation, with concurrent reductions in radiation exposure and operator fatigue \cite{yao2025advancing, stevenson2022robotic, fry2023maintenance}. An overview of the expert-augmented navigation paradigm is illustrated in Fig.~\ref{fig:Introduction}. 

Advancing toward task-level autonomy in this context holds substantial clinical value by promoting consistent navigation across patient-specific anatomies, timely responses to intraoperative changes, and reduced cognitive burden on clinicians \cite{li2026hierarchical, pore2023autonomous, yao2025sim2real}. Reinforcement learning (RL) has emerged as a promising foundation by enabling agents to learn navigation strategies through environmental interaction \cite{dupont2025grand,yan2026mobile}, with growing evidence of transferability to physical systems \cite{yao2025sim4endor, karstensen2025learning}. Despite this momentum, current approaches struggle with low sample efficiency, dependence on hand-crafted rewards, and limited adaptability to evolving intraoperative conditions \cite{cho2022sim}. For example, DDPG in CathSim improved over PPO but relied on uninformed random exploration and remained simulation-only \cite{tian2023ddpg, jianu2024cathsim}. Inverse RL from offline demonstrations can infer rewards yet may miss real-time expert strategies and clinical constraints \cite{robertshaw2024autonomous}. Planning-driven systems that pair DQN with fuzzy control assist drilling but depend on static preoperative models, limiting responsiveness to imaging dynamics and precluding online policy refinement \cite{ji2023heuristically}. Critically, autonomous catheter steering for bifurcation navigation remains underexplored, with few methods explicitly targeting real-time steering decisions that integrate live imaging, online expert feedback, and robust control on physical platforms\cite{truesdell2023intravascular,yao2023enhancing,luo2025precise}. These gaps motivate sample-efficient learning with online expert correction for autonomous catheter steering in endovascular navigation.

In this context, catheter steering for bifurcation navigation is an online perception–control problem \cite{yao2025real}. The system must detect approaching bifurcations from live imaging, estimate centerlines and relative tip-to-branch pose, and select the appropriate daughter vessel. It then has to orient and advance the catheter and guidewire through the junction while regulating torque, axial push, and rotation to maintain luminal traversal and avoid excessive wall\cite{yao2025real3D,lunardi2022definitions}. This task is challenging due to 2D imaging ambiguity and foreshortening, deformable tool–vessel interactions with friction and backlash, and patient-specific anatomical variability\cite{yao2025multi}. Sparse terminal rewards at branch entry and delayed credit assignment further reduce sample efficiency, while fixed heuristic rewards limit adaptability to intraoperative changes\cite{peloso2025imitation}. These factors motivate learning strategies that inject expert priors, enable online correction to keep exploration safe and informative, and incorporate execution-time compensation for modeling errors during branch selection.

This paper presents a sample‑efficient learning framework for autonomous catheter steering at vascular bifurcations with online expert correction. The SAC‑EIL‑GAIL framework integrates Soft Actor-Critic (SAC), Generative Adversarial Imitation Learning (GAIL), and expert‑in‑the‑loop supervision to accelerate learning, improve policy fidelity, and preserve adaptability under live imaging. Catheter kinematics are modeled with a constant‑curvature formulation and aligned with image‑derived centerlines using sub‑pixel skeleton fitting. Expert maneuvers are converted into control commands that inform policy updates and define target poses at bifurcations. A fuzzy logic controller compensates for image‑driven modeling errors and stabilizes branch selection. 

The primary contributions of this study are as follows:
\begin{itemize}
    \item A unified SAC-EIL-GAIL framework that combines maximum-entropy RL, adversarial imitation, and online expert correction. The design improves sample efficiency and training stability, enhances policy plausibility through expert guidance, and reduces reliance on hand-crafted rewards.
    \item A constant-curvature kinematic model is coupled with skeleton-based centerline extraction for sub-pixel trajectory fitting. Expert demonstrations are mapped to control inputs, enabling precise steering and reliable localization at challenging bifurcations.
    \item Target poses derived from expert-in-the-loop interaction support bifurcation navigation. A fuzzy logic controller compensates for image-derived modeling errors in real time, improving navigation success rates and robustness in complex vascular environments.
\end{itemize}

\section{Methodology}
\subsection{System Overview}  
The experimental system consists of a robotic catheterization platform and an autonomous navigation framework. As shown in Fig.~\ref{fig:Framework}(a), two paired robotic arms collaborate in operation, where one executes axial insertion together with rotational control of the catheter, and the other provides stabilization to suppress unintended motion and ensure operational safety. Each arm integrates rotary joints and linear stages with grippers that securely manipulate the catheter.  

The software architecture integrates reinforcement learning, imitation learning, and fuzzy control under expert supervision. A YOLO-based vision module continuously detects the catheter tip and vascular bifurcations from intraoperative images. When the tip enters a bifurcation, the intervention module retrieves an appropriate trajectory and posture from a predefined library and refines the motion with fuzzy control. In the remaining segments, actions are autonomously generated by the reinforcement learning agent. Upon reaching the terminal region, the task is rewarded and the catheter is reset to the initial position for the next trial.  

\subsection{Problem Formulation}
The task of autonomous catheter navigation is formulated as a constrained sequential decision-making problem under the reinforcement learning paradigm. The environment is defined by a state space $\mathcal{S}$, an action space $\mathcal{A}$, and the transition distribution $p(s_{t+1}|s_t,a_t)$. At each time step $t$, the catheter system is described by a state $s_t \in \mathcal{S}$, which includes information derived from image segmentation such as the skeletonized centerline, the distal tip position, and its orientation. The agent selects an action $a_t \in \mathcal{A}$, corresponding to base manipulations such as translational advancement or axial rotation, and receives a reward $R_t$ that reflects navigation safety and efficiency. The optimal navigation policy is obtained by maximizing the expected cumulative reward:
\begin{equation}
\pi^* = \arg \max_{\pi} \; \mathbb{E}_{\pi}\left[\sum_{t=0}^{T} R_t \right].
\end{equation}
Physical constraints are imposed based on catheter mechanics, which are represented using a constant curvature approximation with a fixed small-angle distal bend. Under this model, curvature along the catheter centerline $\gamma(s)$ is characterized by  
\begin{equation}
\kappa(s) = \frac{|\gamma'(s) \times \gamma''(s)|}{|\gamma'(s)|^3}, \quad s \in [0,L],
\end{equation}
ensuring geometric stability during navigation while approximating the inherent bending property of the distal segment. During navigation, when the catheter tip reaches a vascular bifurcation, an expert provides corrective guidance by designating a target pose $(D_{\text{target}}, P_{\text{target}})$. The navigation system then minimizes both translational and rotational errors, expressed as $e_{\text{trans}} = \| P_{\text{target}} - P_{\text{current}} \|_2$ and $e_{\text{rot}} = D_{\text{target}} - D_{\text{current}}$, respectively. The learning objective integrates environment-driven reward signals with expert priors through a hybrid reinforcement learning strategy. The overall reward is formulated as 
\begin{equation}
R_t = w_{\mathrm{SAC}}(t)\, r_{\mathrm{SAC}} + w_{\mathrm{GAIL}}(t)\, r_{\mathrm{GAIL}} + \epsilon_t,
\end{equation}
where $\epsilon_t \sim \mathcal{U}(-\delta,\delta)$ introduces stochasticity, and the time-dependent weights $w_{\mathrm{SAC}}(t)$ and $w_{\mathrm{GAIL}}(t)$ provide a smooth transition from exploration dominated by reinforcement signals to demonstration-driven optimization.

\subsection{Catheter Modeling with Online Expert Correction Pose Mapping for Robotic Control}
The catheter navigation process is modeled using an Expert-in-the-Loop (EIL) strategy, where the robotic system continually adjusts its posture by evaluating discrepancies between the current catheter configuration and expert-selected correction poses. These corrections are translated into fuzzy control rules that guide the underlying actuation signals.

\begin{figure*}[!htbp]
\centering
\includegraphics[width=0.96\textwidth]{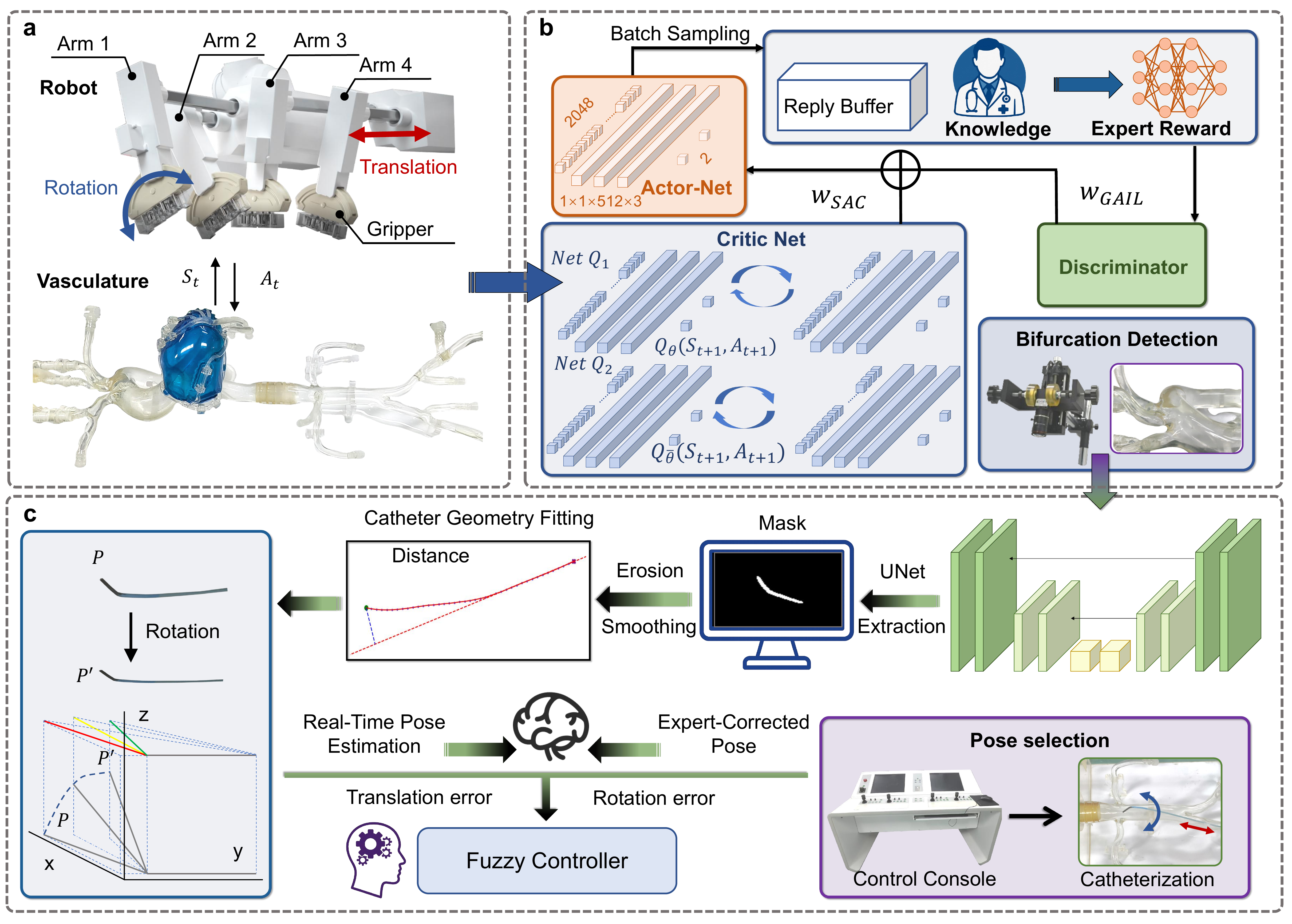}
\caption{Expert-in-the-loop catheter navigation framework integrating reinforcement learning and fuzzy control. (a) Robotic catheterization setup and agent–environment interaction: the agent observes the vascular state $s_t$ and outputs action $a_t$ to command catheter translation, rotation, and gripping. (b) Policy learning that combines Soft Actor–Critic (SAC) with Generative Adversarial Imitation Learning (GAIL): mini-batches from a replay buffer train an actor–critic with twin critics ($Q_1$, $Q_2$). A discriminator supplies an expert reward shaped by prior knowledge, and the policy is optimized using a weighted sum of SAC and GAIL rewards ($w_{\text{SAC}}$, $w_{\text{GAIL}}$). Bifurcation detection triggers the expert-in-the-loop module. (c) Online fuzzy pose correction at bifurcations: an expert selects a target pose; a U-Net-based segmentation extracts the catheter mask, followed by smoothing/erosion and geometry fitting for real-time pose estimation. Translation and rotation errors feed a fuzzy controller that adjusts robot commands to reach the expert-corrected pose.}
\label{fig:Framework}
\end{figure*}

\subsubsection{Skeleton Extraction}
The catheter skeleton is derived from the segmented image by applying a thinning algorithm, where pixels are represented as foreground with value 1 and background with value 0 \cite{van2014scikit}. Each candidate foreground pixel $p_1$ and its eight-neighborhood $\{p_2, \ldots, p_9\}$ are retained as skeletal points only if a set of logical conditions is satisfied: the pixel itself must belong to the foreground, the total number of neighboring pixels must lie between two and six, the connectivity number $S(p_1)$ must be equal to one, and simultaneous occupancy of certain triplets of neighbors must be avoided to prevent spurious branches. The connectivity number is formally defined as
\begin{equation}
S(p_1) = \sum_{i=2}^{9} s(p_i, p_{i+1}),
\end{equation}
where $s(p_i, p_{i+1})$ encodes a 0-to-1 transition along the neighborhood in a clockwise order.

Once a coherent skeletonized structure is obtained, the trajectory extraction is guided by geometric significance. The Euclidean distance transform is employed to identify prominent endpoints, where the pixel associated with the largest distance value is labeled as $Q_1$ and the farthest endpoint relative to $Q_1$ is designated as $Q_2$. The longest path connecting these two points is computed using a breadth-first search procedure, ensuring maximal coverage of the skeletal structure. To improve numerical stability and mitigate discretization artifacts, the extracted path is further smoothed with a Savitzky–Golay filter, yielding a continuous sub-pixel centerline representation suitable for subsequent modeling and control, as illustrated in Fig.~\ref{fig:Framework}(b).

\begin{algorithm}[t]
\caption{Online Expert-Corrected Reinforcement Learning with Fuzzy Control}
\KwIn{Expert-selected pose image $I_{\text{expert}}$, initial state-action pair $(s_{\text{init}}, a_{\text{init}})$, expert dataset $(s_{\text{expert}}, a_{\text{expert}}, r_{\text{expert}})$}
\KwOut{Optimal policy $\pi^*$}
\If{$episodes < 50$}{
    $\pi \leftarrow \text{SoftActorCriticExplore}(s_{\mathrm{SAC}}, a_{\mathrm{SAC}})$; record reward $r_{\mathrm{SAC}}$\;
}
\If{$episodes \geq 50$}{
    $R_t \leftarrow w_{\mathrm{SAC}}(t)\,r_{\mathrm{SAC}} + w_{\mathrm{GAIL}}(t)\,r_{\mathrm{GAIL}} + \epsilon_t$\;
    $\pi^* \leftarrow \text{TrainPolicy}(s_t,a_t,R_t)$\;
    \If{bifurcation detected}{
        $(D_{\text{target}}, P_{\text{target}}) \leftarrow \text{ExtractPose}(I_{\text{expert}})$\;
        $(D_{\text{current}}, P_{\text{current}}) \leftarrow \text{ExtractPose}(I_{\text{current}})$\;
        \While{$(|e_{\text{rot}}| > \epsilon_{\text{rot}})\,\lor\, (|e_{\text{trans}}| > \epsilon_{\text{trans}})$}{
            $e_{\text{rot}} \leftarrow D_{\text{target}} - D_{\text{current}}$\;
            $e_{\text{trans}} \leftarrow \|P_{\text{target}} - P_{\text{current}}\|_2$\;
            $a_{\text{rule}} \leftarrow \text{FuzzyControl}(e_{\text{rot}}, e_{\text{trans}})$\;
            $\text{Update}(D_{\text{current}}, P_{\text{current}})$\;
        }
    }
}
\end{algorithm}

\subsubsection{Control Input Mapping}
The rotation angle to reproduce the expert pose is inferred from the 2D segmented skeleton. A catheter with a fixed distal bend is modeled as a polyline rotating about its central axis. By rotating the base, the manipulator controls the tip orientation and overall pose. As shown in Fig.~\ref{fig:Framework}(b), the catheter tip endpoint \( P(x, y, z) \) is rotated about the x-axis by an angle \( \theta \), resulting in a new position \( P'(x', y', z') \). The projection of \( P' \) onto the x–y plane gives a vertical distance \( d' = \left| y \cos\theta - z \sin\theta \right| \) from the x-axis. When \( \theta_{\text{pitch}} \in \left(0, \frac{\pi}{2}\right) \), \( d' \) decreases monotonically, and when \( \theta_{\text{pitch}} \in \left(\frac{\pi}{2}, \pi\right) \), it increases. This distance \( d' \) represents the distance from \( P' \) to the fitted line of the proximal shaft. The pitch angle \( \theta_{\text{pitch}} \) is mapped to the actuator input \( u_r \) via: $u_r = \mathcal{F}(\theta_{\text{pitch}}) = \mathcal{F}(G(D))$.

In order to distinguish between the possible positive and negative values of \( y' \) after rotation in the \( (0, \pi) \) range, we select two points \( A(0, b) \) and \( B(1, k+b) \) on the line \( kx - y + b = 0 \), as shown in the following formula:
\begin{equation}
D = \begin{cases} 
d' & \text{if } \overrightarrow{AB} \times \overrightarrow{AP} > 0, \\
-d' & \text{if } \overrightarrow{AB} \times \overrightarrow{AP} < 0.
\end{cases}
\end{equation}

Upon the completion of the catheter modeling, a fuzzy control approach is employed to map the robot's control inputs to the actions selected by the expert. The control algorithm computes the translation and rotation errors from image modeling, adjusting the position and orientation to reach the target. The translation error \( e_{\text{trans}} =  \| (P_{\text{target}} , P_{\text{current}}) \|_2\) is the Euclidean distance between the target point and the current catheter tip , and the rotation error is \( e_{\text{rot}} =  D_{\text{target}}-D_{\text{current}}\).
\subsubsection{Fuzzy Control based on Expert Correction}
The input error \( e \) is mapped to predefined fuzzy categories (NL, NS, Z, PS, PL) with centers \( m \). A triangular membership function, parameterized by a constant half-width \( w \), calculates the degree of membership for each fuzzy set, effectively transforming the crisp input into linguistic variables with associated membership values.
\begin{equation}
\mu(e) = 
\begin{cases}
0, & e \leq m - w \text{ or } e \geq m + w \\
\frac{e - (m - w)}{w}, & m - w < e \leq m \\
\frac{(m + w) - e}{w}, & m < e < m + w
\end{cases}
\end{equation}

Fuzzy inference is performed using a rule base defined over all combinations of the fuzzified inputs. Each rule's activation strength is calculated using the minimum operator (representing fuzzy AND): 

\begin{equation}
    \mu_{\text{rule}} = \min\left( \mu_t(e_{\text{trans}}), \mu_r(e_{\text{rot}}) \right).
\end{equation}

The control outputs are aggregated using the maximum operator (fuzzy OR) across all activated rules: 

\begin{equation}
    \mu_{C}(u) = \max\left( \mu_{C}(u), \mu_{\text{rule}} \right).
\end{equation}

The final crisp control outputs are obtained using the centroid method:
\begin{equation}
u^* = \frac{\sum_{i} \mu_i \cdot m_i}{\sum_{i} \mu_i}
\end{equation}
where \(\mu_i\) and $m_i$denote the membership degree and center of the $i$-th output fuzzy set.
\subsection{Expert Experience Guided Trajectory Optimization}
We propose a hybrid reinforcement learning framework that integrates SAC with Generative Adversarial Imitation Learning (GAIL). The method gradually shifts the reward contribution from environment-driven signals to expert imitation, enabling early exploration and later incorporation of expert priors for balanced learning.
\subsubsection{Exploration-Oriented Reinforcement Learning with Soft Actor–Critic Algorithm}
SAC is an off-policy deep reinforcement learning method based on maximum entropy policy optimization\cite{haarnoja2018soft}. It aims to maximize the expected cumulative reward while encouraging higher entropy in the policy to promote exploration. Its objective function is formulated as:
\begin{equation}
J_{\pi} = \sum_{t=0}^{T} \mathbb{E}_{(s_t,a_t) \sim \rho_{\pi}} \left[ r(s_t,a_t) + \alpha \mathcal{H}(\pi(\cdot|s_t)) \right]
\end{equation}
where $r(s_t,a_t)$ denotes the instantaneous environmental reward, $\mathcal{H}(\pi(\cdot|s_t))$ is the policy entropy, and $\alpha$ is the temperature coefficient.
\subsubsection{Expert-Guided Policy Shaping via Generative Adversarial Imitation Learning}
GAIL aims to minimize the divergence between the agent's trajectory distribution and that of an expert via adversarial training \cite{ho2016generative}. A discriminator $D_{\psi}$ is trained to differentiate between expert trajectories and agent-generated ones, while the policy network $\pi_{\theta}$ attempts to fool the discriminator:
\begin{equation}
\begin{aligned}
    \hspace{-1em}
\min_{\pi_{\theta}} \max_{D_{\psi}} \ \mathbb{E}_{\pi_{\theta}}\left[ \log \left( 1 - D_{\psi}(s,a) \right) \right] \\
+ \mathbb{E}_{\pi_E} \left[ \log D_{\psi}(s,a) \right]
\end{aligned}
\end{equation}

The imitation reward from the discriminator is given by:
\begin{equation}
r_{\mathrm{GAIL}}(s,a) = -\log\left( 1 - D_{\psi}(s,a) \right)
\end{equation}

This transforms expert prior knowledge into a reward signal that guides the agent toward expert-like behavior.
\subsubsection{Hybrid Reward Scheduling for Balancing Exploration and Imitation}
To achieve a smooth transition from reinforcement learning to imitation learning, we propose a sigmoid-based dynamic reward scheduling mechanism. During training, rewards from the SAC and GAIL branches are combined with time-dependent weights, enabling a gradual shift from exploration to demonstration-guided behavior.Let $T$ be the total number of training episodes and $t$ the current episode. The weighting factor is defined as:

\begin{table*}[t]
\centering
\caption{Comparison of Autonomous Navigation Across RL Algorithms for Bifurcation Steering}
\label{tab:rl_comparison}
\setlength{\tabcolsep}{6pt}
\renewcommand{\arraystretch}{1.3}
\small
\begin{tabular*}{\textwidth}{@{\extracolsep{\fill}} l c c c c c}
\toprule
\textbf{Algorithm}  & \textbf{Episodes} & \textbf{Avg. Steps} & \textbf{Success Rate} & \textbf{Avg. Time (s)} & \textbf{Avg. Error (px)} \\
\midrule
TD3            & 300 & 9.21 & 41.67\% (125/300) & 78.09 & 208.76 \\
SAC            & 300 & 8.32 & 44.67\% (134/300) & 60.39 & 98.55 \\
SAC-GAIL       & 300 & 7.57 & 53.67\% (161/300) & 59.04 & 149.53 \\
SAC-EIL        & 300 & 7.66 & 51.33\% (154/300) & 61.05 & 86.90 \\
SAC-EIL-GAIL   & 300 & 7.61 & 59.00\% (177/300) & 59.41 & 82.58 \\
\bottomrule
\end{tabular*}
\vspace{0.5ex}
\begin{minipage}{\textwidth}
\footnotesize
Note: An episode is counted as successful if it reaches the target and is flanked by five consecutive successful episodes both before and after. Avg. Time denotes the average duration per episode until convergence. Avg. Error is the mean deviation from the expert pose (translation + rotation) at bifurcations, computed per successful episode.
\end{minipage}
\end{table*}

\begin{figure*}[!htbp]
\centering
\includegraphics[width=1.0\textwidth]{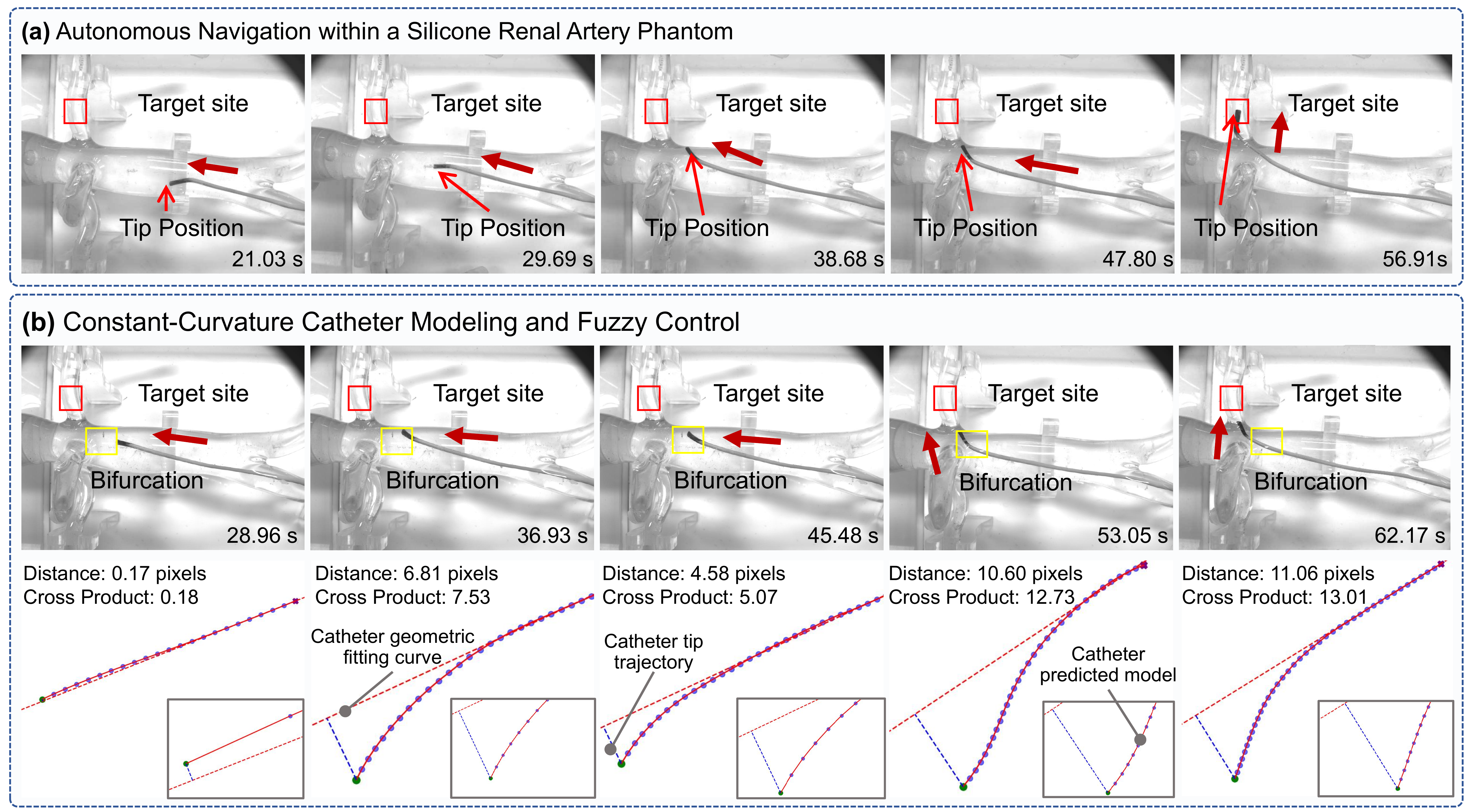}
\caption{Autonomous catheter navigation and constant-curvature modeling. (a) After training, the learned policy autonomously steers the catheter to the predefined target region in a silicone renal artery phantom by continuously localizing the tip in fluoroscopic frames. (b) Upon bifurcation detection, the catheter centerline (skeleton) is extracted and fitted with a constant-curvature model. Directional consistency is assessed by the sign/magnitude of the cross product between the fitted tangent and the skeleton tangent; the resulting geometric errors drive an expert-in-the-loop fuzzy controller that updates robot commands to achieve the expert-specified pose. Insets report the tip–target distance (pixels) and cross-product values over time.}
\label{fig:Result}
\end{figure*}

\begin{equation}
\alpha(t) = \frac{1}{1 + \exp\left( -k \cdot \left(t - \frac{T}{2} \right) \right)}
\end{equation}
\begin{equation}
w_{\mathrm{SAC}}(t) = 1 - 0.5 \cdot \alpha(t), \quad w_{\mathrm{GAIL}}(t) = 0.5 \cdot \alpha(t)
\end{equation}

where $k$ is a temperature parameter that controls the steepness of the transition. As training progresses, $w_{\mathrm{SAC}}(t)$ gradually decreases from 1 to 0.5 while $w_{\mathrm{GAIL}}(t)$ increases from 0 to 0.5, ensuring a smooth shift from exploration to exploitation.

The total reward at episode $t$ is then defined as:
\begin{equation}
R_t = w_{\mathrm{SAC}}(t) \cdot r_{\mathrm{SAC}} + w_{\mathrm{GAIL}}(t) \cdot r_{\mathrm{GAIL}} + \epsilon_t
\end{equation}

where $\epsilon_t \sim \mathcal{U}(-\delta, \delta)$ is a uniform random perturbation that encourages stochasticity and robustness during learning.

By replacing the original reward in SAC’s Bellman objective with $R_t$, the optimization becomes:
\begin{equation}
J_{\pi} = \sum_{t=0}^{T} \mathbb{E} \left[ R_t + \alpha \mathcal{H}(\pi(\cdot|s_t)) \right].
\end{equation}

\section{Experiments and Results}
\subsection{Implementation Details}
Experiments were conducted on a 3D silicone renal artery phantom filled with saline, which provided a controlled vascular environment. Catheter manipulation was performed using a pair of custom-built 2-DoF robotic arms: one executed translational and rotational motions to mimic physician operation, while the other provided system stabilization. The entire setup was monitored by a Hikvision MV-CS050-10GM industrial camera at 10 frames per second and computations were accelerated on an NVIDIA RTX 3090 GPU.

Catheter tip detection relied on a YOLOv5 model trained on 7,293 brightness-augmented images \cite{jocher2020ultralytics}. After each action, the centroid position $P(x,y)$ was extracted, with detection failures ($x < 0$) prompting a short pause and reacquisition. When consecutive failures occurred, the last valid frame was reused to maintain temporal consistency. Catheter segmentation was performed using a U-Net model trained on 1,000 augmented images derived from 400 originals. The tip was identified by fitting a line through 30\% of one endpoint and selecting the opposing end.

\begin{figure}[t]
\centering
\includegraphics[width=0.48\textwidth]{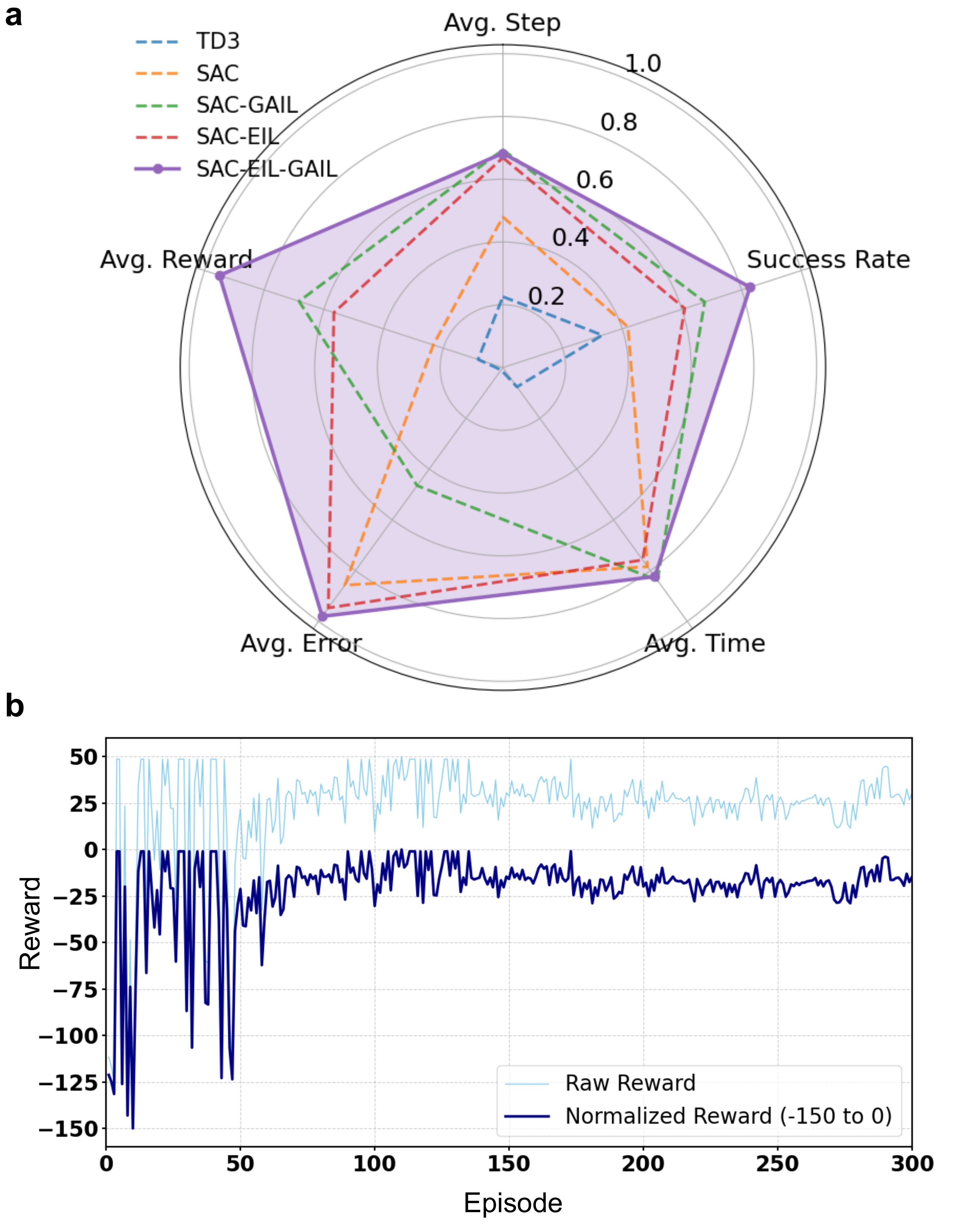}
\caption{Comparison of algorithm performance metrics. (a) Normalized Comparison of Key Performance Indicators Across Different RL Algorithms. (b) Average reward per episode during \text{SAC-EIL-GAIL} training. 
As the number of episodes increases, the reward gradually converges. 
This indicates stable and effective learning.}
\label{fig:Result_figure1}
\end{figure}

Fuzzy control employed five linguistic sets \{NL, NS, Z, PS, PL\} for both translation and rotation. Translation membership centers were \{-2.0, -1.0, 0, 1.0, 2.0\}\,cm with half-width $\Delta t = 1.0\,\text{cm}$, derived from the calibrated mapping $D \in [0,215]\,\text{pixel} \to d \in [0,2.5]\,\text{cm}$. Rotation membership centers were $-60^\circ, -30^\circ, 0^\circ, 30^\circ, 60^\circ$ with half-width $\Delta r = 30^\circ$, based on the mapping $E \in [0,80] \to e \in [0,90^\circ]$, where $e = (80 - E) \times 1.125^\circ/\text{pixel}$.

The reward function combined sparse terminal signals with distance-based shaping. Episodes terminated with a penalty of $r_t = -150$ when push distance exceeded 500\,mm, step count exceeded 20, or the catheter moved out of bounds ($x_h < 10$ or $x_h > 900$). In all other cases, $r_t$ was shaped by the distance to the target, with additional penalties for rotational deviation and misalignment.

\subsection{Model Performance Evaluation}
The navigation task is meticulously designed to evaluate both the accuracy and efficiency of autonomous guidance within a 3D anatomical model. As illustrated in Fig.\ref{fig:Result}(a), navigation initiates at the femoral artery entry point in the lower right and proceeds autonomously toward the target region in the upper left, guided by real-time coordinates of the catheter tip. To assess navigation performance, five reinforcement learning (RL) frameworks were trained and evaluated across 300 episodes summarized in Table \ref{tab:rl_comparison}. The baseline algorithm was implemented via PyTorch and Stable-Baselines3\cite{raffin2021stable}.
\begin{figure}[t]
\centering
\includegraphics[width=0.48\textwidth]{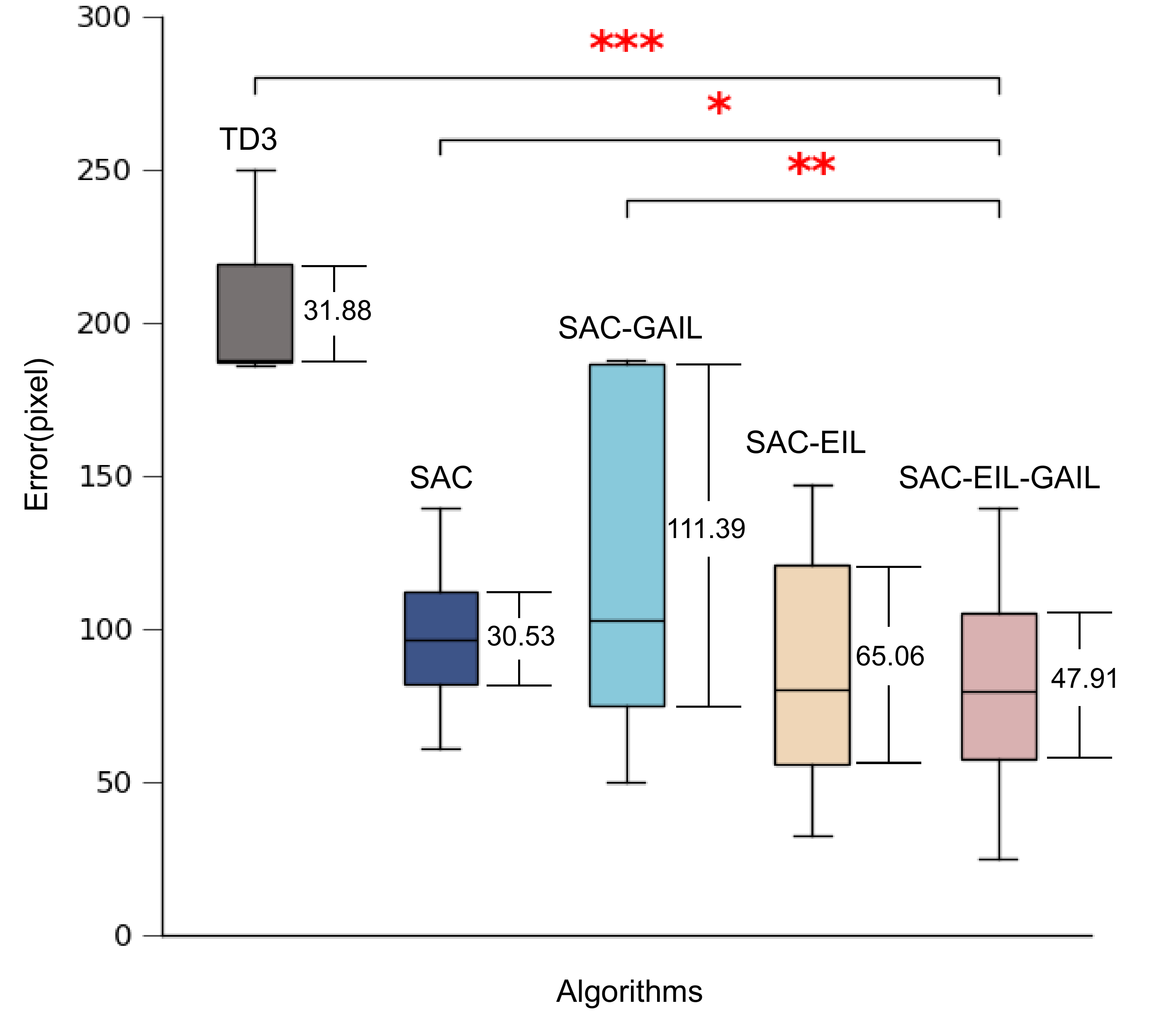}
\caption{This figure illustrates the error distributions of TD3, SAC, SAC-GAIL, SAC-EIL, and SAC-EIL-GAIL. The boxplots highlight the central tendency and variability of each method, while the red asterisks indicate statistically significant differences with SAC-EIL-GAIL ($*p < 0.05, **p < 0.01, ***p < 0.001$).}
\label{fig:Result_figure2}
\end{figure}

The proposed method exhibits clear over the baseline algorithms TD3 and SAC across multiple evaluation criteria. With respect to convergence, within 300 episodes, the proposed approach attains stable convergence after only 123 episodes, whereas TD3 and SAC require 175 and 166 episodes, respectively. Since fewer episodes indicate faster convergence, this result substantiates the enhanced convergence efficiency of the proposed method. Moreover, the convergence time is shortened to 59.41 seconds, further demonstrating superior execution performance. 

The ablation study demonstrates that integrating expert imitation learning with adversarial learning significantly improves policy quality. The full model achieves a success rate of 59.00\%, which is 5.33 percentage points higher than SAC-GAIL and 7.67 points higher than SAC-EIL, confirming its superior task reliability. The updated SAC-EIL-GAIL algorithm has an average step count of 7.61, slightly higher than SAC-GAIL’s 7.57 with an increase of approximately 0.53\%, but lower than SAC-EIL’s 7.66 with a reduction of about 0.65\%. The slight increase in steps may result from the adaptation process after incorporating expert data, as the policy requires some extra steps to adjust. This is a common phenomenon in imitation-reinforcement learning integration. Although the average episode duration shows marginal differences, 1.64 seconds faster than SAC-EIL and 0.37 seconds slower than SAC-GAIL, the full model demonstrates an overall improvement in efficiency. 

These results validate the proposed method’s superiority in both baseline and ablation evaluations, particularly in enhancing policy reliability and efficiency through the combination of expert imitation and adversarial learning. The consistent improvements across multiple metrics suggest that the integration of these two components not only accelerates learning but also fosters more stable and adaptable strategies. By leveraging expert demonstrations to guide early exploration and incorporating adversarial mechanisms to refine long-term decision-making, the method establishes a complementary progression that enhances both learning stability and adaptability.

\subsection{Error-Based Analysis and Confidence Evaluation}
The normalized comparison of algorithmic performance across key metrics, as illustrated in Fig.\ref{fig:Result_figure1}(a), highlights the overall advantage of the proposed SAC-EIL-GAIL framework.  SAC-EIL-GAIL achieves the lowest average error of 82.58 pixels, significantly smaller than TD3 (208.76), SAC (98.55), SAC-GAIL (149.53), and SAC-EIL (86.90). This reduction in deviation reflects the improved precision of the learned navigation policy, particularly in handling complex bifurcation scenarios.

The error distribution achieved by SAC-EIL-GAIL exhibits a clear advantage over those of TD3, SAC, and SAC-GAIL, with lower variance and tighter concentration around the minimal error range, as shown in Fig.~\ref{fig:Result_figure2}. Statistical comparisons, with p-values falling below conventional significance thresholds, support the robustness of this improvement. The boxplot underscores SAC-EIL-GAIL’s consistent and precise performance, particularly under more demanding conditions. These results suggest that integrating expert imitation learning with adversarial regularization not only enhances convergence efficiency but also improves trajectory accuracy, yielding a more reliable framework for autonomous navigation.

\section{Discussion}
The SAC-EIL-GAIL framework consistently outperforms both baseline and ablation methods, achieving faster convergence, higher success rates, and the lowest navigation error. Statistical analysis confirms that combining online expert correction with adversarial imitation enhances policy quality and yields stable improvements across all key metrics. This hybrid design enables the agent to adapt rapidly while leveraging expert knowledge for high-precision navigation, particularly in complex bifurcation scenarios.  

Comparison with ablation settings shows that, although both SAC-EIL-GAIL and SAC-EIL employ online correction, the relative advantage in step efficiency over SAC-GAIL is less pronounced. This arises from the additional adjustments required during the early training phase, when adaptation to expert guidance introduces more refinements. As training progresses, the initial alignment improves and fewer corrections are needed, indicating that efficiency gains become most apparent near convergence, while early-stage benefits lie primarily in stability.  

In terms of accuracy, SAC-EIL-GAIL achieves the lowest error and mitigates the limitations of SAC in reward stability, ensuring faster convergence and more reliable performance. Nonetheless, runtime improvements remain modest and the advantage over SAC-EIL is incremental. Occasional fluctuations in error further suggest that robustness under varied conditions is not yet fully achieved. These results highlight that SAC-EIL-GAIL represents a meaningful advance in autonomous catheter navigation but also indicate the necessity for further refinement to enhance robustness, efficiency, and generalization under diverse clinical environments.  

\section{Conclusion and Future Work}
This paper proposed SAC-EIL-GAIL, an imitation learning framework that combines expert demonstrations and prior knowledge with reinforcement learning to enhance robot autonomy in navigating vascular bifurcations. By leveraging expert data and training on a silicone renal artery phantom, the framework improves pose adjustment when entering branch vessels, shortens training time, and achieves a 17.33\% higher success rate compared to the TD3 baseline.

Future directions will focus on multi-instrument collaborative navigation of guidewires and catheters, given their high flexibility and complex nonlinear behavior. Another direction will focus on adaptive expert feedback mechanisms that can operate online, providing corrective inputs during training and execution without compromising safety. Additional emphasis will be placed on enhancing generalization through simulation-to-real transfer and patient-specific vascular modeling, enabling the framework to accommodate diverse anatomical geometries and clinical conditions. In the longer term, integration into the clinical workflow has the potential to support patient-specific rehearsal, intraoperative guidance, and postoperative assessment, thereby contributing to safer, more precise, and more efficient robot-assisted endovascular interventions.

\bibliographystyle{ieeetr}
\balance
\bibliography{reference}

@article{li2024machine,
  title={Machine learning to predict outcomes of endovascular intervention for patients with PAD},
  author={Li, Ben and Warren, Blair E and Eisenberg, Naomi and Beaton, Derek and Lee, Douglas S and Aljabri, Badr and Verma, Raj and Wijeysundera, Duminda N and Rotstein, Ori D and de Mestral, Charles and others},
  journal={JAMA Network Open},
  volume={7},
  number={3},
  pages={e242350--e242350},
  year={2024},
  publisher={American Medical Association}
}

@article{konda2025robotically,
  title={Robotically steerable guidewires—Current trends and future directions},
  author={Konda, Revanth and Brumfiel, Timothy A and Bercu, Zachary L and Grossberg, Jonathan A and Desai, Jaydev P},
  journal={Science Robotics},
  volume={10},
  number={105},
  pages={eadt7461},
  year={2025},
  publisher={American Association for the Advancement of Science}
}

@article{yan2026mobile,
  title={A Mobile Magnetic Manipulation Platform for Gastrointestinal Navigation with Deep Reinforcement Learning Control},
  author={Yan, Zhifan and Liu, Chang and Jiang, Yiyang and Zheng, Wenxuan and Chen, Xinhao and Krieger, Axel},
  journal={arXiv preprint arXiv:2601.15545},
  year={2026}
}

@article{yao2025advancing,
  author={Yao, Tianliang and Lu, Bo and Kowarschik, Markus and Yuan, Yixuan and Zhao, Hubin and Ourselin, Sebastien and Althoefer, Kaspar and Ge, Junbo and Qi, Peng},
  journal={IEEE Reviews in Biomedical Engineering}, 
  title={Advancing Embodied Intelligence in Robotic-Assisted Endovascular Procedures: A Systematic Review of AI Solutions}, 
  year={2026},
  volume={19},
  number={},
  pages={248-266}
}

@article{stevenson2022robotic,
  title={Robotic-assisted PCI: the future of coronary intervention?},
  author={Stevenson, Alexander and Kirresh, Ali and Ahmad, Mahmood and Candilio, Luciano},
  journal={Cardiovascular Revascularization Medicine},
  volume={35},
  pages={161--168},
  year={2022},
  publisher={Elsevier}
}

@misc{fry2023maintenance,
  title={Maintenance of competence in cardiovascular practice: it’s time for more learning, less testing},
  author={Fry, Edward TA and Kuvin, Jeffrey and Sibley, Janice},
  journal={Journal of the American College of Cardiology},
  volume={81},
  number={9},
  pages={924--927},
  year={2023},
  publisher={American College of Cardiology Foundation Washington DC}
}

@article{pore2023autonomous,
  title={Autonomous navigation for robot-assisted intraluminal and endovascular procedures: A systematic review},
  author={Pore, Ameya and Li, Zhen and Dall'Alba, Diego and Hernansanz, Albert and De Momi, Elena and Menciassi, Arianna and Gelpi, Alicia Casals and Dankelman, Jenny and Fiorini, Paolo and Vander Poorten, Emmanuel},
  journal={IEEE Transactions on Robotics},
  volume={39},
  number={4},
  pages={2529--2548},
  year={2023},
  publisher={IEEE}
}

@article{yao2025sim2real,
  title={Sim2real learning with domain randomization for autonomous guidewire navigation in robotic-assisted endovascular procedures},
  author={Yao, Tianliang and Wang, Haoyu and Lu, Bo and Ge, Jiajia and Pei, Zhiqiang and Kowarschik, Markus and Sun, Lining and Seneviratne, Lakmal and Qi, Peng},
  journal={IEEE Transactions on Automation Science and Engineering},
  year={2025},
  volume={22},
  number={},
  pages={13842-13854},
  publisher={IEEE}
}

@article{dupont2025grand,
  title={The grand challenges of learning medical robot autonomy},
  author={Dupont, Pierre E and Degirmenci, Alperen},
  journal={Science Robotics},
  volume={10},
  number={104},
  pages={eadz8279},
  year={2025},
  publisher={American Association for the Advancement of Science}
}

@INPROCEEDINGS{yao2025sim4endor,
  author={Yao, Tianliang and Ban, Madaoji and Lu, Bo and Pei, Zhiqiang and Qi, Peng},
  booktitle={2025 IEEE International Conference on Robotics and Automation (ICRA)}, 
  title={Sim4EndoR: A Reinforcement Learning Centered Simulation Platform for Task Automation of Endovascular Robotics}, 
  year={2025},
  volume={},
  number={},
  pages={824-830},
}

@article{karstensen2025learning,
  title={Learning-based autonomous navigation, benchmark environments and simulation framework for endovascular interventions},
  author={Karstensen, Lennart and Robertshaw, Harry and Hatzl, Johannes and Jackson, Benjamin and Langej{\"u}rgen, Jens and Breininger, Katharina and Uhl, Christian and Sadati, SM Hadi and Booth, Thomas and Bergeles, Christos and others},
  journal={Computers in Biology and Medicine},
  volume={196},
  pages={110844},
  year={2025},
  publisher={Elsevier}
}

@article{peloso2025imitation,
  title={Imitation Learning for Path Planning in Cardiac Percutaneous Interventions},
  author={Peloso, Angela and Damiano, Rossella and Zhang, Xiu and Bicchi, Anna and Votta, Emiliano and De Momi, Elena},
  journal={IEEE Transactions on Biomedical Engineering},
  year={2025},
  publisher={IEEE}
}

@inproceedings{tian2023ddpg,
  title={A DDPG-based method of autonomous catheter navigation in virtual environment},
  author={Tian, Wei and Guo, Jian and Guo, Shuxiang and Fu, Qiang},
  booktitle={2023 IEEE International Conference on Mechatronics and Automation (ICMA)},
  pages={889--893},
  year={2023},
  organization={IEEE}
}

@article{jianu2024cathsim,
  title={CathSim: an open-source simulator for endovascular intervention},
  author={Jianu, Tudor and Huang, Baoru and Vu, Minh Nhat and Abdelaziz, Mohamed EMK and Fichera, Sebastiano and Lee, Chun-Yi and Berthet-Rayne, Pierre and y Baena, Ferdinando Rodriguez and Nguyen, Anh},
  journal={IEEE Transactions on Medical Robotics and Bionics},
  volume={6},
  number={3},
  pages={971--979},
  year={2024},
  publisher={IEEE}
}

@article{robertshaw2024autonomous,
  title={Autonomous navigation of catheters and guidewires in mechanical thrombectomy using inverse reinforcement learning},
  author={Robertshaw, Harry and Karstensen, Lennart and Jackson, Benjamin and Granados, Alejandro and Booth, Thomas C},
  journal={International Journal of Computer Assisted Radiology and Surgery},
  volume={19},
  number={8},
  pages={1569--1578},
  year={2024},
  publisher={Springer}
}

@article{luo2025precise,
  title={Precise and dexterous robotic manipulation via human-in-the-loop reinforcement learning},
  author={Luo, Jianlan and Xu, Charles and Wu, Jeffrey and Levine, Sergey},
  journal={Science Robotics},
  volume={10},
  number={105},
  pages={eads5033},
  year={2025},
  publisher={American Association for the Advancement of Science}
}

@article{ji2023heuristically,
  title={A heuristically accelerated reinforcement learning-based neurosurgical path planner},
  author={Ji, Guanglin and Gao, Qian and Zhang, Tianwei and Cao, Lin and Sun, Zhenglong},
  journal={Cyborg and Bionic Systems},
  volume={4},
  pages={0026},
  year={2023},
  publisher={AAAS}
}

@article{truesdell2023intravascular,
  title={Intravascular imaging during percutaneous coronary intervention: JACC state-of-the-art review},
  author={Truesdell, Alexander G and Alasnag, Mirvat A and Kaul, Prashant and Rab, Syed Tanveer and Riley, Robert F and Young, Michael N and Batchelor, Wayne B and Maehara, Akiko and Welt, Frederick G and Kirtane, Ajay J and others},
  journal={Journal of the American College of Cardiology},
  volume={81},
  number={6},
  pages={590--605},
  year={2023},
  publisher={American College of Cardiology Foundation Washington DC}
}

@article{yao2023enhancing,
  title={Enhancing percutaneous coronary intervention with heuristic path planning and deep-learning-based vascular segmentation},
  author={Yao, Tianliang and Wang, Chengjia and Wang, Xinyi and Li, Xiang and Jiang, Zhaolei and Qi, Peng},
  journal={Computers in Biology and Medicine},
  volume={166},
  pages={107540},
  year={2023},
  publisher={Elsevier}
}

@article{yao2025real,
  title={Real-Time Guidewire Tip Tracking Using a Siamese Network for Image-Guided Endovascular Procedures},
  author={Yao, Tianliang and Pei, Zhiqiang and Li, Yong and Yuan, Yixuan and Qi, Peng},
  journal={Advanced Intelligent Systems},
  pages={2500425},
  year={2025},
  publisher={Wiley Online Library}
}

@article{lunardi2022definitions,
  title={Definitions and standardized endpoints for treatment of coronary bifurcations},
  author={Lunardi, Mattia and Louvard, Yves and Lef{\`e}vre, Thierry and Stankovic, Goran and Burzotta, Francesco and Kassab, Ghassan S and Lassen, Jens F and Darremont, Olivier and Garg, Scot and Koo, Bon-Kwon and others},
  journal={Journal of the American College of Cardiology},
  volume={80},
  number={1},
  pages={63--88},
  year={2022},
  publisher={American College of Cardiology Foundation Washington DC}
}

@inproceedings{yao2025real3D,
  title={Real-Time 3D Guidewire Reconstruction from Intraoperative DSA Images for Robot-Assisted Endovascular Interventions},
  author={Yao, Tianliang and Li, Bingrui and Lu, Bo and Pei, Zhiqiang and Yuan, Yixuan and Qi, Peng},
  booktitle={2025 IEEE/RSJ International Conference on Intelligent Robots and Systems (IROS)},
  pages={17344--17351},
  year={2025},
  organization={IEEE}
}

@article{yao2025multi,
  author={Yao, Tianliang and Xu, Yueqi and Wang, Haoyu and Qiu, Xihe and Althoefer, Kaspar and Qi, Pen},
  journal={IEEE Transactions on Fuzzy Systems}, 
  title={Multi-Agent Fuzzy Reinforcement Learning With LLM for Cooperative Navigation of Endovascular Robotics}, 
  year={2025},
  volume={},
  number={},
  pages={1-11}
}

@inproceedings{cho2022sim,
  title={Sim-to-real transfer of image-based autonomous guidewire navigation trained by deep deterministic policy gradient with behavior cloning for fast learning},
  author={Cho, Yongjun and Park, Jae-Hyeon and Choi, Jaesoon and Chang, Dong Eui},
  booktitle={2022 IEEE/RSJ International Conference on Intelligent Robots and Systems (IROS)},
  pages={3468--3475},
  year={2022},
  organization={IEEE}
}

@inproceedings{haarnoja2018soft,
  title={Soft actor-critic: Off-policy maximum entropy deep reinforcement learning with a stochastic actor},
  author={Haarnoja, Tuomas and Zhou, Aurick and Abbeel, Pieter and Levine, Sergey},
  booktitle={International Conference on Machine Learning},
  pages={1861--1870},
  year={2018},
  organization={Pmlr}
}

@article{ho2016generative,
  title={Generative adversarial imitation learning},
  author={Ho, Jonathan and Ermon, Stefano},
  journal={Advances in neural information processing systems},
  volume={29},
  year={2016}

}

@article{raffin2021stable,
  title   = {Stable-Baselines3: Reliable Reinforcement Learning Implementations},
  author  = {Raffin, Antonin and Hill, Ashley and Gleave, Adam and Kanervisto, Anssi and Ernestus, Matthias and Dormann, Noah},
  journal = {Journal of Machine Learning Research},
  volume  = {22},
  number  = {268},
  pages   = {1--8},
  year    = {2021},
  url     = {http://jmlr.org/papers/v22/20-1364.html}
}

@article{jocher2020ultralytics,
  title={ultralytics/yolov5: v3. 0},
  author={Jocher, Glenn and Stoken, Alex and Borovec, Jirka and Changyu, Liu and Hogan, Adam and Diaconu, Laurentiu and Poznanski, Jake and Yu, Lijun and Rai, Prashant and Ferriday, Russ and others},
  journal={Zenodo},
  year={2020}
}

@article{van2014scikit,
  title={scikit-image: image processing in Python},
  author={Van der Walt, Stefan and Sch{\"o}nberger, Johannes L and Nunez-Iglesias, Juan and Boulogne, Fran{\c{c}}ois and Warner, Joshua D and Yager, Neil and Gouillart, Emmanuelle and Yu, Tony},
  journal={PeerJ},
  volume={2},
  pages={e453},
  year={2014},
  publisher={PeerJ Inc.}
}

@article{li2026hierarchical,
  author={Li, Yamei and Ge, Ruijian and Zhu, Aoji and Zhao, Jiachi and Shi, Danjing and Sun, Yinghan and Li, Yangmin and Yang, Lidong},
  journal={IEEE Robotics and Automation Letters}, 
  title={A Hierarchical Framework for Real-Time Path Planning of Microswarm in Dynamic Environments}, 
  year={2026},
  volume={11},
  number={3},
  pages={3891-3898}
}
\end{document}